\documentclass{article}

\usepackage{arxiv}

\usepackage[utf8]{inputenc} 
\usepackage[T1]{fontenc}    
\usepackage{hyperref}       
\usepackage{url}            
\usepackage{booktabs}       
\usepackage{amsfonts}       
\usepackage{nicefrac}       
\usepackage{microtype}      

\usepackage{amsmath}
\usepackage{multirow}
\usepackage{array}
\usepackage{multicol}
\usepackage{comment}
\usepackage{graphicx}
\usepackage{subfigure}
\usepackage[center]{caption}
\usepackage{balance}
\usepackage[noadjust]{cite}

\title{Predicting Helpfulness of Online Reviews}

\author{
  Abdalraheem Alsmadi\\
  Computer Science Department\\
  Jordan University of Science and Technology\\
  Irbid, Jordan \\
  \texttt{afalsmadi16@cit.just.edu.jo} \\
   \And
 Shadi AlZu'bi \\
  Computer Science Department\\
  Al Zaytoonah University of Jordan\\
  Amman, Jordan \\
  \texttt{smalzubi@zuj.edu.jo} \\
  \AND
  Mahmoud Al-Ayyoub\\
  Computer Science Department\\
  Jordan University of Science and Technology \\
  Irbid, Jordan \\
  \texttt{maalshbool@just.edu.jo} \\
  \And
  Yaser Jararweh \\
  Computer Science Department\\
  Jordan University of Science and Technology \\
  Irbid, Jordan \\
  \texttt{yijararweh@just.edu.jo} \\
}

\begin{document}
\maketitle

\begin{abstract}
E-commerce dominates a large part of the world's economy with many websites dedicated to online selling products. The vast majority of e-commerce websites provide their customers with the ability to express their opinions about the products/services they purchase. These feedback in the form of reviews represent a rich source of information about the users' experiences and level of satisfaction, which is of great benefit to both the producer and the consumer. However, not all of these reviews are helpful/useful. The traditional way of determining the helpfulness of a review is through the feedback from human users. However, such a method does not necessarily cover all reviews. Moreover, it has many issues like bias, high cost, etc. Thus, there is a need to automate this process. This paper presents a set of machine learning (ML) models to predict the helpfulness online reviews. Mainly, three approaches are used: a supervised learning approach (using ML as well as deep learning (DL) models), a semi-supervised approach (that combines DL models with word embeddings), and pre-trained word embedding models that uses transfer learning (TL). The latter two approaches are among the unique aspects of this paper as they follow the recent trend of utilizing unlabeled text. The results show that the proposed DL approaches have superiority over the traditional existing ones. Moreover, the semi-supervised has a remarkable performance compared with the other ones.
\end{abstract}


\section{Introduction}
\label{sec:intro}

Online reviews of different product/services might be considered as one of the most important references for many customers. The vast majority of websites that are interested in selling products online allow people who have already made the purchase to express their opinion about the bought product or service by writing online reviews. At the same time, some of these sites allow new customers and site visitors to vote on each of these reviews as a helpful review or not. As a result, new customers can benefit from such voting to decide which reviews to read/consider when making their decision to product/service or not~\cite{duandynamics}.

As reading a plenty of such reviews of a certain product/service and knowing the helpful ones from the others may be not easy for a new user~\cite{opinion2017}, studying and evaluating these reviews are very important, as minimized list of only the helpful reviews are provided to the new customers. Furthermore, it is also important to differentiate between real ones or fake ones. It is necessary to have an efficient method to automatically find whether the review is helpful or not~\cite{alsmadi2019using}.

In this research, we focus on identification of helpful reviews using a set of machine/deep learning methods following three approaches: supervised approach, semi-supervised approach, and pre-trained model. The semi-supervised approach is among the unique aspects of the proposed work and it is motivated by recent works~\cite{RCNN_Lai,ioHRCNN} that showed that supplementing supervised learning algorithms trained on a labeled dataset with features and/or models learned from unsupervised data can lead to improved performance. All approaches are trained and tested using a subset of Amazon reviews dataset~\cite{dataset2} prepared by~\cite{SmadiBrief}. As for the methods we adopt in the supervised approach, we used a set of classifiers including FastText~\cite{fasttext}, Support-Vector-Machine (SVM), unidirectional and multilayer bidirectional Recurrent Neural Network (RNN), Convolutional Neural Network (CNN) based on Kim implementation in~\cite{CNN_senClasi_kim2014convolutional}, and Recurrent Convolutional Neural Network (RCNN) based on the proposed architecture by Lai et al.~\cite{RCNN_Lai}. As for the semi-supervised approach, we adopt only the RCNN model. As for the pre-trained models, we use the BERT~\cite{Bert} and RoBERTa~\cite{Roberta} models. All mentioned methods in all approaches have been evaluated using the accuracy measure.

The remaining parts of this paper is structured as follows. The background and related literature are discussed in Section~\ref{sec:Background and Related Work}. A description of the dataset and how it is prepared is provided in Section~\ref{sec:Datset}.
Section~\ref{sec:Experimentation} presents the proposed models and their results. Section~\ref{sec:Conclusion} concludes the proposed work and the future expected work is provided.

\section{Related Works}
\label{sec:Background and Related Work}
In this part, we give a succinct background on the problem at hand and discuss the previous efforts addressing it.

As we mentioned before, any new customer is concerned about reviews and previous customers experience about a specific product. Reviews are well employed in increasing the products sales percentage, where product evaluation is used to connect the purchase of products with its reviews~\cite{duandynamics}. Many factors are considered to explain the reviewers’ evaluation of each separate product, where some of them are to direct reviewers’ fulfilment of that product and vice versa, or if the product meets minimum requirements or not. Product delivery speed, real product characteristics, and advertisement match is considered as well.

\begin{figure*}
\centering
\includegraphics[width= 14cm, height = 5cm]{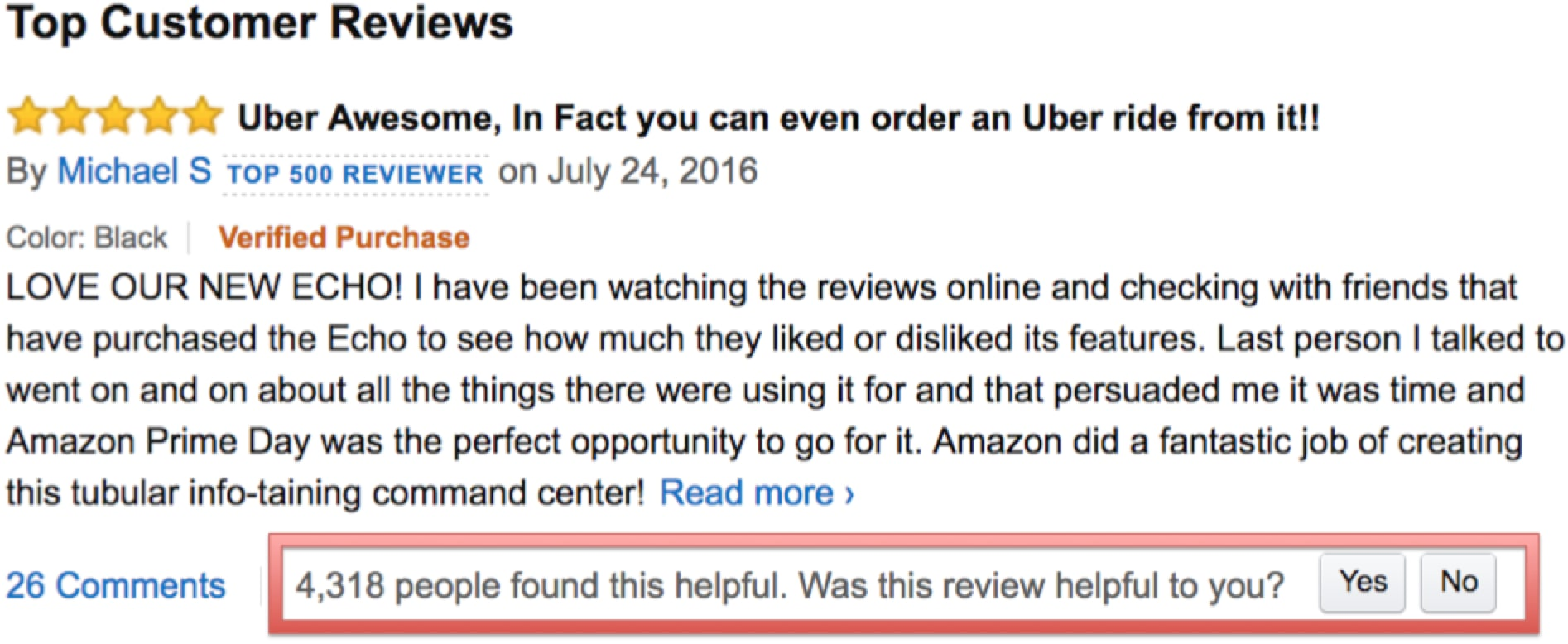}
\caption{How Amazon review voting works.}
\label{fig1_example}
\end{figure*}

Figure~\ref{fig1_example} shows a sample review and how the customers can vote for its helpfulness~\cite{PredictingAmazonProductReview-Wei}. Moreover, we show below two sample reviews of contrasting helpfulness scores:
\begin{itemize}
\item{\textbf{category: Movies \& TV}}\\
Reviewer Text: ``Not too bad at the beginning, but you expect it to build to something more than it does.  Kind of a let down with the ending.'', ``helpful'': [1,9], ``overall'': 2.0.
\item{\textbf{category: CDs \& Vinyl}}\\
Reviewer Text: ``I've never felt such sadness for something that happened so long ago because of music!  I felt my Scottish blood stirring as I listened to how my ancestors were forced from their homes.  All Scotsmen (and women) should have this CD as part of their collection.'', ``helpful'': [4,4], ``overall'': 5.0.
\end{itemize}
From the cited reviews, we noticed that both texts delimited in each cited review, and we can notice the number of votes for each review. However, in the Movies \& TV category review, it can be noticed that one of nine persons has recommended this review as a helpful and voted up here, while the others did not find the review helpful. Similarly, in the CDs \& Vinyl category review, we noticed the whole four persons have recommended this review as a helpful and voted up here. 
Reading the first review sensibly, you can judge that no helpful content available. This is the explanation of the unhelpful review eight votes. The opposite can be observed about the second review.

The Amazon products' reviews dataset has garnered a lot of attention as it provides a huge amount of information about product purchases, customer reviews, and various metadata for purchases from the website. As the focus of this study is to predict the helpfulness of reviews, this subsection is limited to the related works focused only on this problem, especially the ones employing Deep Learning (DL) models. In a similar work, by measuring the influential users like~\cite{measuringtwitter,mypaper,aishapaper}. In~\cite{mypaper}, Horani et al. built-up a co-purchase network to identify and extract the influential users and how they are differentiated from other users using the relationship between centrality measures and user's reviewing records.

In~\cite{EvaluateHelpfulnessReport}, Nguy defined this as a binary classification problem, he proposed Recurrent Neural Network (RNN) and Long Short Term Memory (LSTM) cells to solve it. Three categories were used to extract the reviews from slight subsets named ``5-cores'' (``Digital Music'', ``Musical Instruments'', ``Patio, Lawn and Garden''), it have been extracted from the Amazon dataset. The tested dataset have 88,239 reviews, which includes > 50\% reviews with positive votes and considered helpful. The proposed model by Nguy accomplished an accuracy of 65\% and F1 measure of 54.9\%.

Wei et al.~\cite{PredictingAmazonProductReview-Wei} proposed an RNN model for helpfulness prediction. The resources were limited. Therefore, they chosed 50,000 random reviews from the ``5-core'' dataset. The dataset was divided into two parts (training  80\% and testing 20\%). The authors used an RNN model, which recognized good reviews (accuracy; 80.5\%, area under the curve (AUC); 88\%), and bad reviews (accuracy: 75.7\%, AUC: 83\%).

Qu et al.~\cite{ReviewHelpfulnessAssessmentbasedonConvolutionalNeuralNetwork} proposed a CNN model for reviews helpfulness assessment. The authors compared two approaches for the initialization of word vectors: random and using Global Vectors (Glove)~\cite{glove} with diverse review according to the length of words. Two main classes were considered in the experiments: ``Books'' and ``Electronics'', where reviews with small number of votes was excluded. Similar to~\cite{EvaluateHelpfulnessReport}, if the helpfulness votes percentage of any product is more than 50\%, it would be considered as helpful. Their model achieved accuracy (75.17\%) and F1 measure (75\%) for the Books category. While in Electronics category accuracy (76.96\%) and F1 measure (77\%).

\section{Dataset Description and Preparation}
\label{sec:Datset}
This section is dedicated to describe the dataset in details and discuss the process taken to prepare it for this work.

\subsection{Description and Analysis}
In this work, we use the 2014 version of the Amazon reviews dataset~\cite{dataset2}, which is publicly available.\footnote{\url{http://jmcauley.ucsd.edu/data/amazon/}}
The dataset consists of 83.7 million exclusive reviews from 24 main product categories straddling between MAY-1996 and JUL-2014. The top 4 categories (which we consider in this work) are illustrated in Table~\ref{tab1}, as well as the reviews and products in each category.

\begin{table}
\centering
\caption{Amazon dataset description.}
\label{tab1}
\begin{tabular}{@{}|c|c|c|@{}}
\toprule
\multicolumn{3}{|c|}{Amazon reviews dataset}\\ \midrule
\textbf{Category} & \textbf{Contained Reviews} & \textbf{Covered Products} \\ \midrule
Movies and TV & 4,607,047 &208,321 \\ \midrule
CDs and Vinyl& 3,749,004 &492,799 \\ \bottomrule
Electronics   & 7,824,482  & 498,196 \\ \midrule
Books&22,507,155  & 2,370,585 \\ \midrule
\end{tabular}
\end{table}

In the dataset, every single record provides one review for each product in the corresponding category with the following tags:
\begin{itemize}
\item {\textbf{reviewerID}: }the reviewer ID on Amazon.
\item {\textbf{asin}: }the unique identification number of the item.
\item{\textbf{reviewerName}: }includes the reviewer name.
\item{\textbf{helpful}: }the rate of the review according to helpfulness, such $[20/32]$.
\item{\textbf{reviewText}: }the review text body.
\item{\textbf{overall}: }the product rating value.
\item{\textbf{summary}: }the review summary.
\item{\textbf{unixReviewTime}: }Unix time of the review.
\item{\textbf{reviewTime}: }the review date and time.
\end{itemize}

\begin{figure}
\centering
\includegraphics[width=0.47\textwidth, height = 4.5cm]{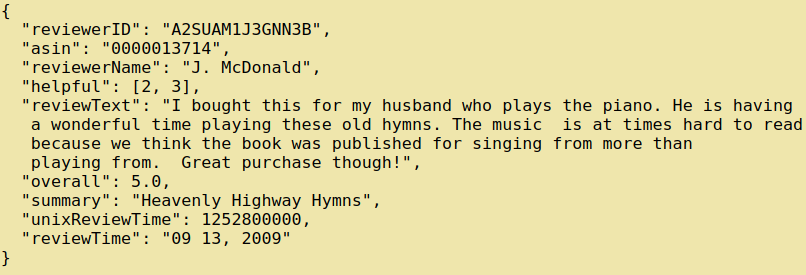}
\caption{Amazon dataset: A review sample.}
\label{reviewSample}
\end{figure}

A review sample of the employed dataset is illustrated in Figure~\ref{reviewSample}~\cite{SmadiBrief}.

\begin{figure*}
\centerline{
\subfigure[Books category]{
 \includegraphics[width=3.5in]{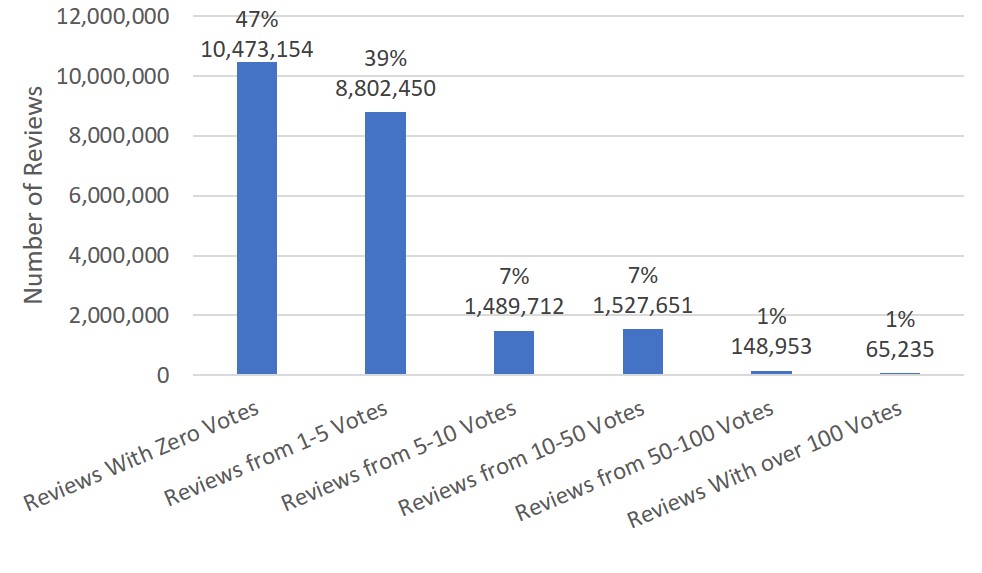}
 \label{fig:Books}
}
\subfigure[Electronics category]{
 \includegraphics[width=3.5in]{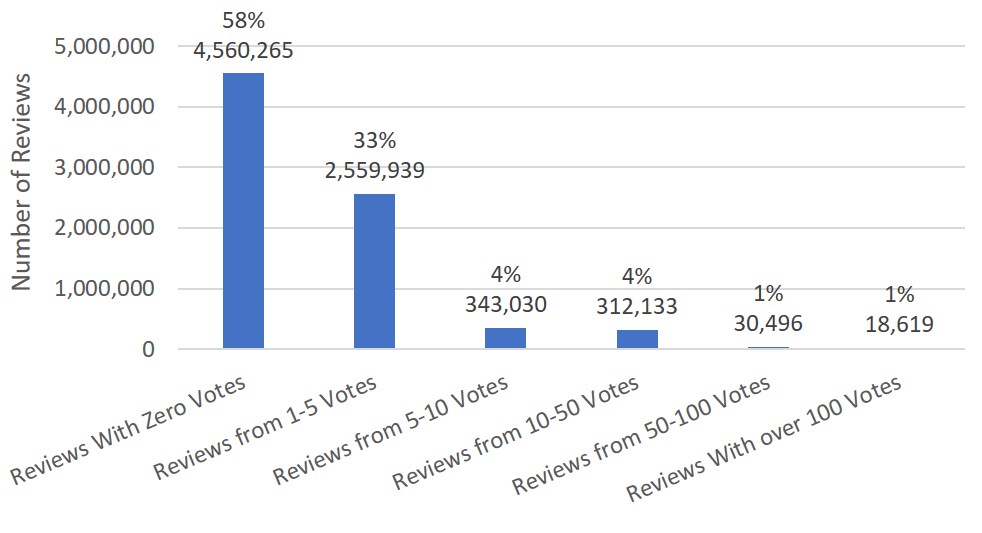}
 \label{fig:Electronics}
}
}
\centerline{
\subfigure[Movies and TV category]{
 \includegraphics[width=3.5in]{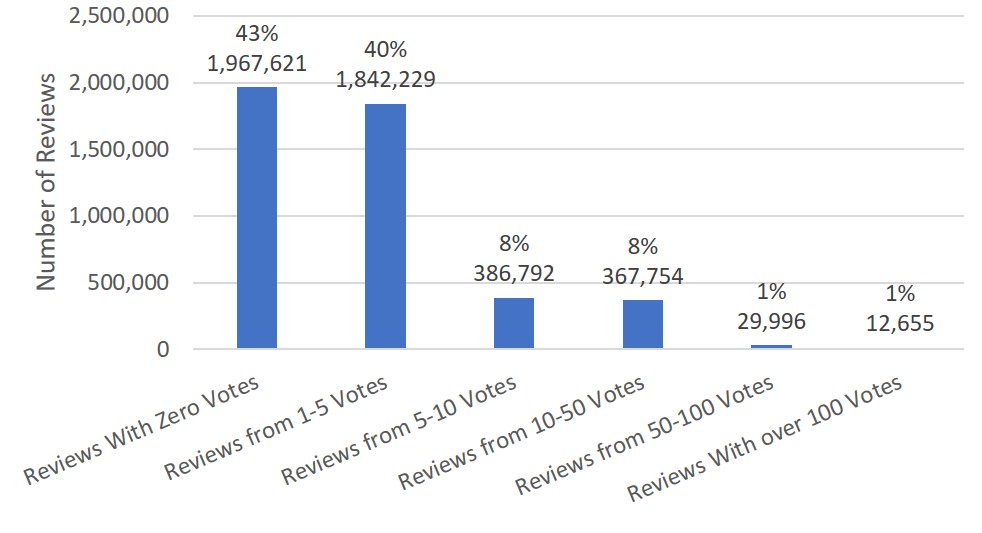}
 \label{fig:Movies_and_TV}
}
\subfigure[CDs and Vinyl category]{
 \includegraphics[width=3.5in]{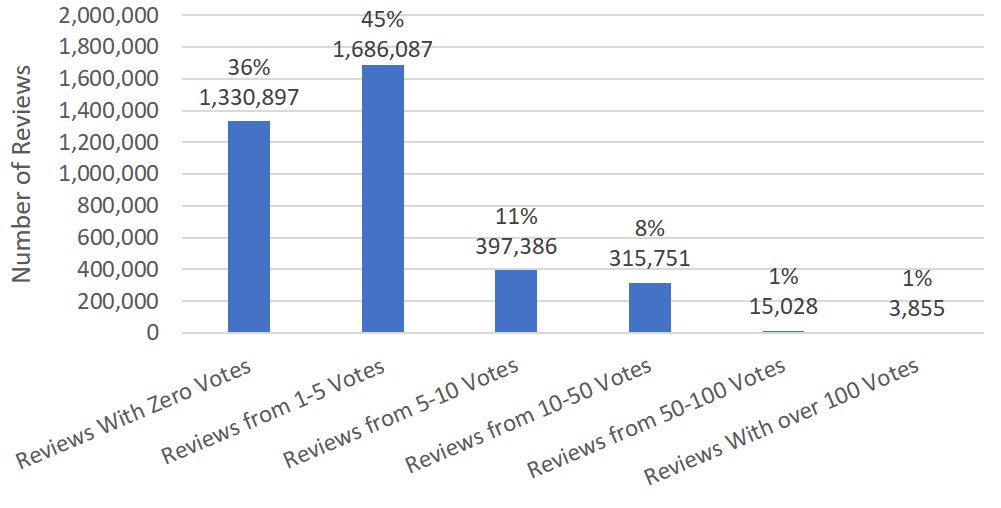}
 \label{fig:CDs_and_Vinyl}
}
}
\caption{Voting distribution for the four categories.}
\label{fig:voting_dist}
\end{figure*}

The work presented in ~\cite{SmadiBrief}, overviews a comprehensive statistical analysis on the same dataset used in this work. One of the aspects covered in that analysis was the distribution of the reviews based on the number of votes they received into six classes:
0 votes,
1-5 votes,
5-10 votes,
10-50 votes,
50-100 votes, and
more than 100 votes.
Figure~\ref{fig:voting_dist} shows these voting distributions for the four categories under consideration (Taken from ~\cite{SmadiBrief}).
From this figure, we can observe that about half of the reviews received no votes.
One might be tempted to assume that such reviews are unhelpful ones since many voters follow the quote: ``if you don't have anything nice to say, don't say anything at all''.
However, after manual inspection of the dataset, we realized that such reviews need not be unhelpful or contain no useful information. In fact, some of them are very insightful and informative. However, they may have failed to grab the attention of the voters for many reasons such as the lack of interest in the product itself at the period between writing the review and collecting the data.

In addition to reviews with zero votes, there is a large percentage (39\%) of reviews with very small number of votes (1-5 votes). This means that prior works that simply excluded reviews with less than six reviews practically lost 89\% of the data, which might contain a wealth of information. Our goal is to explore these untapped resources to improve the helpfulness prediction models performance.
On the other hand, one can observe a sharp drop in the number of reviews that had a large number of votes. Such power-law probability distribution is very common in social networks. In both cases, we are interested in studying these two phenomena and taking advantage of this analysis.

\subsection{Text Preparation and Normalization}
A research was conducted by Ghose et al. for defining and determining helpful reviews~\cite{ghose2011estimating}, the authors compared several experts views regarding to the review helpfulness. They assign 60\% from all review votes as an edge to consider it as helpful. After careful inspection of the data in the the work~\cite{SmadiBrief,Employing}, we took a more extreme approach in assigning helpful/unhelpful labels to the reviews. Specifically, we consider a review helpful only if the percentage it received is more than 75\% of the overall votes to ensure that the helpful reviews are indeed helpful without any doubt. One the other hand, if the percentage of the helpful votes it received is less than 35\%, then unhelpful review is assigned here.

One point to mention here is that the Amazon dataset is published as JSON files in which the reviews are sorted by the reviewer name. Therefore, to ensure that the extracted reviews for the proposed experiments cover the majority of items in each category, we shuffled each one of the four files randomly many times. For the reviews with non-zero votes, we extracted 100,000 reviews from each category, where half of the reviews are helpful and the other half is unhelpful according to the definitions in the previous paragraph. We also made sure that the maximum review length is 500 words and that each review has at least ten votes. As for the zero-vote reviews, we extracted 400,000 reviews from each category with maximum review length of 500 words/review.

\section{Proposed Models and Evaluation}
\label{sec:Experimentation}
\begin{figure*}
\centering
\includegraphics[width= 0.90 \textwidth, height = 7 cm]{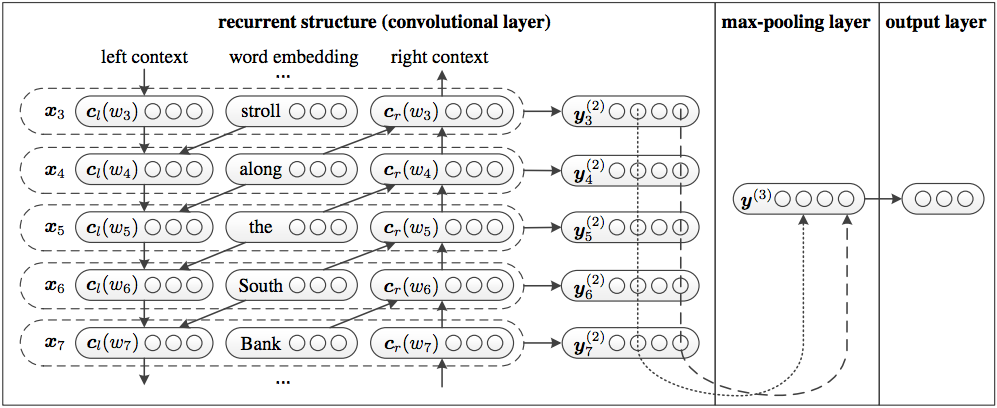}
\caption{RCNN structure.}
\label{fig: rcnn}
\end{figure*}

This work aims at predicting the helpfulness of reviews in three ways. The first one, which we call Task 1 (T1), is to approach this problem in a supervised manner. The second one, which we call Task 2 (T2), is to approach this problem in a semi-supervised manner. And the third one which we call Task 3 (T3), is to approach this problem using pre-trained models.
To handle the tasks (T1) and (T2), we develop a model based on the RCNN architecture proposed by Lai et al.~\cite{RCNN_Lai}. The implementation uses the TensorFlow platform. As its name suggests, RCNN is a combination between RNN and CNN. The strength of this model lies in combining the advantages of the two models as follows. It uses Bidirectional RNN (Bi-RNN)~\cite{bidirectional}, which is suitable for contextual information capturing. As for the CNN part, its benefit comes from the ``max-pooling'' with its ability to determine the keywords in the text that are playing a major role in determining its class. Figure~\ref{fig: rcnn} represents the architecture of the RCNN model used~\cite{RCNN_Lai}. For the third task (T3) we used the BERT~\cite{Bert} and RoBERTa~\cite{Roberta} models in the experiments to predict the helpful reviews using the same subset from task 1 (T1).

\subsection{Proposed Approaches}
\subsubsection{\textbf{Supervised Approach}}
As we mentioned before, the T1 dataset consists of 100,000 labeled reviews per category: 50,000 helpful reviews and 50,000 unhelpful reviews. We shuffle the dataset randomly and divided it into 2 sets: training and validation (90\%) and testing (10\%). The first part for training purposes results in 90,000 reviews, evenly divided, 45,000 helpful reviews, and 45,000 unhelpful reviews. The second part, which is for testing purposes, resulted in 10,000 reviews evenly divided, 5,000 as a helpful review, and 5,000 as an unhelpful review. From the training set part, 15\% has been taken for validation and hyper-parameter tuning, which results in 13,500 reviews.

In order to address T1, a model based on RCNN is developed. The model uses a Bi-RNN and it is fed the whole text and all terms existing. Word embedding features are computed using random uniform initialization and embedding vector size of 256. Adam optimizer is used for optimization with a learning-rate of 0.001.

\subsubsection{\textbf{Semi-Supervised Approach}}
As we mentioned earlier, about half of the dataset is unlabeled, and the unique aspect of the proposed work is to benefit from this by approaching the prediction of the helpfulness reviews in a semi-supervised manner. To do so, 400,000 unlabeled reviews/category are extracted from the dataset representing the unsupervised part in this experiment. They are merged with the training set from the supervised approach (T1), which has 90,000 labeled reviews/category. Hence, the total number of reviews/category for the training part becomes 490,000 (labeled and unlabeled) reviews. The testing set (10,000 labeled reviews/category) from (T1) is kept the same. The total number of reviews for (T2) in the experiments is 500,000 reviews/category, which is an unprecedentedly massive number.

For T2, the same approach of T1 is developed based on RCNN trained along with a pre-trained word embedding model. FastText~\cite{fasttext} is used to initialize the word look-up table for the word embeddings based on the skip-gram model~\cite{20mikolov2013efficient} on both labeled and unlabeled training dataset. Adam optimizer is used for optimization with a learning-rate of 0.001. An important modification on the RCNN routine is detected after using FastText based word embedding.
Table~\ref{tab:T2parameters} shows the main parameters setup used in RCNN model for both approaches (T1 \& T2) experiments.

\begin{table}
\centering
\caption{Parameters setup for (T1 \& T2) experiments.}
\label{tab:T2parameters}
\begin{tabular}{@{}|c|l|c|c|@{}}
\toprule
\multirow{2}{*}{\#} & \multirow{2}{*}{\textbf{parameter}} & \multicolumn{2}{c|}{Value}\\ \cmidrule(l){3-4}
&& Task 1 (T1)& Task 2 (T2)\\ \midrule
1.& Number of classes& 2& 2\\ \midrule
2.& Batch size& 128& 128\\ \midrule
3.& Number of epochs& 10& 10\\ \midrule
4.& Words max length& 500& 500\\ \midrule
5.& Word embedding & Random Uniform & Fasttext
(Skip-gram) \\ \midrule
6.& Embedding size& 256& 300\\ \midrule
7.& RNN hidden units& 256& 300\\ \midrule
8.& FC hidden units& 256& 300\\ \midrule
9.& Optimizer& Adam& Adam\\ \midrule
10.& Learning rate& 1e-3& 1e-3\\ \bottomrule
\end{tabular}
\end{table}

\begin{table*}
\centering
\caption{Parameters setup for (T3) experiments.}
\label{tab:T3parameters}
\begin{tabular}{@{}|c|c|c|c|c|c|c|@{}}
\toprule
Category                                             & Model   & Training Time & Number of Epochs & Sequence length      & Learning Rate & Number of Classes  \\ \midrule
\multirow{2}{*}{Books}                               & BERT    & 7 Hours       & 2                & \multirow{2}{*}{500} & 2e-5          & \multirow{2}{*}{2} \\ \cmidrule(lr){2-4} \cmidrule(lr){6-6}
                                                     & RoBERTa & 4 Hours       & 2                &                      & 1e-3          &                    \\ \midrule
\multirow{2}{*}{Electronics}                         & BERT    & 7 Hours       & 2                & \multirow{2}{*}{500} & 2e-5          & \multirow{2}{*}{2} \\ \cmidrule(lr){2-4} \cmidrule(lr){6-6}
                                                     & RoBERTa & 7 Hours       & 3                &                      & 1e-3          &                    \\ \midrule
\multicolumn{1}{|l|}{\multirow{2}{*}{CDs and Vinyl}} & BERT    & 7 Hours       & 2                & \multirow{2}{*}{500} & 2e-5          & \multirow{2}{*}{2} \\ \cmidrule(lr){2-4} \cmidrule(lr){6-6}
\multicolumn{1}{|l|}{}                               & RoBERTa & 4 Hours       & 2                &                      & 1e-3          &                    \\ \midrule
\multicolumn{1}{|l|}{\multirow{2}{*}{Movies and TV}} & BERT    & 7 Hours       & 2                & \multirow{2}{*}{500} & 2e-5          & \multirow{2}{*}{2} \\ \cmidrule(lr){2-4} \cmidrule(lr){6-6}
\multicolumn{1}{|l|}{}                               & RoBERTa & 4 Hours       & 2                &                      & 1e-3          &                    \\ \bottomrule
\end{tabular}
\end{table*}
\subsubsection{\textbf{Pre-trained Models}}
Recently, pre-trained language models have shown a significant role to improve many NLP tasks performance, like sentence classification, question answering, and natural language inference~\cite{aisha2019}.
\begin{itemize}
\item \textbf{BERT}: This model~\cite{Bert} relies on the multi-head self-attention mechanism, which enables it to achieve the state-of-the-art accuracy on a wide range of tasks such as, natural language inference, question answering, and sentence classification. The architecture of BERT model is built upon the transformer layer, which is called the self-attention layer. For each layer, the representations of words are exchanged from previous layers regardless of their positions, in contrast to traditional unidirectional models. For each input word, the model learns bidirectional encoder representations by using the masked language model, which randomly masks some of the words from the input to predict the masked word contextually~\cite{Bert}. As BERT offers pre-trained models for English language and multilingual model for 104 different languages, we applied sentence classification through fine-tuning the English language model. Figure~\ref{fig: bert} represents the architecture of the BERT model~\cite{bertfig}.

\begin{figure}
\centering
\includegraphics[width= 0.5 \textwidth]{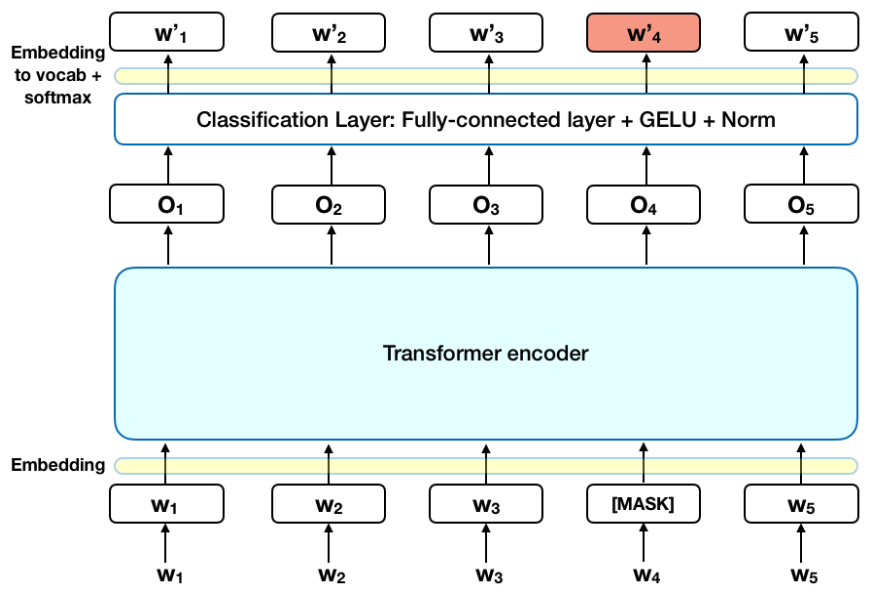}
\caption{BERT structure}
\label{fig: bert}
\end{figure}

\item \textbf{RoBERTa}: RoBERTa~\cite{Roberta} is a replication model of BERT, which relies on a careful evaluation of the effects of hyperparameter tuning and training set size, that can match or exceed the performance of all of the post-BERT methods. The major difference from BERT were in training the model longer, extending data and batch sizes, removing the next sentence prediction objective, training on longer sentences, and dynamically changing the masking pattern applied to the training data.
\end{itemize}
Table~\ref{tab:T3parameters} shows the main parameters setup used in (T3) experiments.

\subsection{Results and Findings}
In addition to RCNN model, several classifiers have been evaluated only on the supervised part, such as: FastText-based, SVM, Bidirectional-LSTM with three layers, and CNN. A set of experiments are conducted to fine-tune each of these classifiers and Table~\ref{tab:supervisedresultsNN} shows the best results of these experiment in addition to the RCNN results. The table shows that the considered models are all outperformed by the supervised RCNN model. Moreover, these numbers are relatively higher than what is reported in the literature on datasets extracted from the Amazon dataset. Unfortunately, none of these datasets is publicly available (to the best of our knowledge), which makes any fair and accurate comparison difficult.
\begin{table}
\centering
\caption{Supervised learning approach results using set of classifiers including RCNN model.}
\label{tab:supervisedresultsNN}
\begin{tabular}{@{}|l|c|c|c|c|c|@{}}
\toprule
\multirow{2}{*}{Category} & \multicolumn{5}{c|}{Accuracy}                                                                            \\ \cmidrule(l){2-6} 
                          & Fasttext & SVM     & \begin{tabular}[c]{@{}c@{}}Bi-LSTM\\  (3 Layers)\end{tabular} & CNN      & RCNN     \\ \midrule
Books                     & 81.4\%  & 75.5\% & 80.9\%                                                       & 78.7\%  & 82\%    \\ \midrule
Electronics               & 80.1\%  & 73.3\% & 80\%                                                         & 77.2\%  & 81\%    \\ \midrule
CDs and Vinyl             & 84.8\%  & 78.5\% & 86\%                                                         & 81.6\%  & 86\%    \\ \midrule
Movies and TV             & 83.9\%  & 76.4\% & 82\%                                                         & 79\%    & 84\%    \\ \bottomrule
Overall                   & 82.55\% & 75.9\% & 82.22\%                                                      & 79.12\% & 83.25\% \\ \bottomrule
\end{tabular}
\end{table}
The results of the semi-supervised RCNN model are shown in Table~\ref{tab:Comparison}.
The table also compares the semi-supervised RCNN results with the supervised RCNN results. It can be observed that the semi-supervised model outperforms the supervised one on all categories. These gains in accuracy are due to the FastText embeddings~\cite{fasttext} with skip-gram model~\cite{20mikolov2013efficient}. The percentage of the improvement on the ``Books'' and ``Electronics'' categories is 5\%, and, for the ``CDs and Vinyl'' and ``Movies and TV's'' categories, it is 6\%.

\begin{table}
\caption{Comparison between experiments accuracies results for RCNN approach.}
\label{tab:Comparison}
\centering
\begin{tabular}{@{}|l|c|c|c|c|c|@{}}
\toprule
\multirow{2}{*}{Category} & \multicolumn{2}{c|}{Training} & \multirow{2}{*}{Testing} & \multicolumn{2}{c|}{Accuracy} \\ \cmidrule(lr){2-3} \cmidrule(l){5-6} 
                          & T1            & T2            &                          & T1            & T2            \\ \midrule
Books                     & 90,000        & 490,000       & 10,000                   & 82\%          & 87\%          \\ \midrule
Electronics               & 90,000        & 490,000       & 10,000                   & 81\%          & 86\%          \\ \midrule
CDs and Vinyl             & 90,000        & 490,000       & 10,000                   & 86\%          & 92\%          \\ \midrule
Movies and TV's           & 90,000        & 490,000       & 10,000                   & 84\%          & 90\%          \\ \bottomrule
\multicolumn{4}{|c|}{Overall}                                                        & 83.25\%       & 88.75\%       \\ \bottomrule
\end{tabular}
\end{table}
The results of the pre-trained models are shown in Table\ref{tab:T3results} in addition to the best results in the supervised approach (T1) and the semi-supervised approach (T2). The results show that working on the pre-trained models like BERT and RoBERTa significantly improved the accuracy compered with previous results, especially the BERT model outperforms all supervised classifiers and RCNN (T1) supervised approach.
\begin{table}
\centering
\caption{The overall best results for (T1) \& (T2) \& Pre-trained models.}
\label{tab:T3results}
\begin{tabular}{@{}|l|c|c|c|c|@{}}
\toprule
\multicolumn{1}{|l|}{\multirow{2}{*}{Category}} & \multicolumn{4}{c|}{Accuracy}                                                                                                    \\ \cmidrule(l){2-5} 
\multicolumn{1}{|c|}{}                          & \begin{tabular}[c]{@{}c@{}}RCNN\\  (T1)\end{tabular} & \begin{tabular}[c]{@{}c@{}}RCNN\\ (T2)\end{tabular} & BERT     & RoBERTa  \\ \midrule
Books                                           & 82 \%                                                & 87 \%                                               & 86.3 \%  & 83.6 \%  \\ \midrule
Electronics                                     & 81 \%                                                & 86 \%                                               & 85.16 \% & 83.8 \%  \\ \midrule
CDs and Vinyl                                   & 86 \%                                                & 92 \%                                               & 87.7 \%  & 85.7 \%  \\ \midrule
Movies and TV's                                 & 84 \%                                                & 90 \%                                               & 85.5 \%  & 83.4 \%  \\ \bottomrule
\multicolumn{1}{|c|}{Overall}                   & 83.25 \%                                             & 88.75 \%                                            & 86.16 \% & 84.12 \% \\ \bottomrule
\end{tabular}
\end{table}

\section{Conclusion and Future Work}
\label{sec:Conclusion}
The area of predicting helpfulness reviews has employed significantly in the past years, and has been considered as many researchers' interest due to its significance and various applications. One common characteristic of the literature on this problem is the huge amount of data that is considered unlabeled, and are thus discarded. The assumption that this data is useless does not always hold as careful inspection proves that some of this data can indeed be useful. In this work, we bride this gap by approaching this problem in three ways: using supervised approach,  semi-supervised approach and using pre-trained models. Specifically for (T1) and (T2), we employed a DL model known as RCNN, which can work as a supervised learning model and a semi-supervised learning model. And for (T3) we employed a two pre-trained models known as BERT and RoBERTa. We first show that the supervised RCNN model is a very powerful one as it outperformed conventional as well as DL-based models. We then show the advantage of the semi-supervised RCNN model over the supervised one. We prove the power of pre-trained models in text classification and how they could outperform the traditional classifiers and DL-based models. For the future work, we are planning to enhance the proposed approaches by applying other DL models and exploring more cutting edge techniques that could improve the prediction and classification results and targeting more categories from the dataset.

\section*{Acknowledgement}
The authors would like to thank the Deanship of Research
at the Jordan University of Science and Technology for supporting
this work (Grant \#20180193).

\bibliographystyle{unsrt}  

\end{document}